\pdfoutput=1

\documentclass[11pt]{article}

\usepackage[colorlinks, linkcolor=blue, citecolor=blue, urlcolor=blue]{hyperref}
\usepackage{float}
\usepackage{listings}

\usepackage{EMNLP2023}

\usepackage{times}
\usepackage{latexsym}

\usepackage[T1]{fontenc}

\usepackage[utf8]{inputenc}
\usepackage{booktabs, multirow, multicol}
\usepackage[usestackEOL]{stackengine}
\usepackage{graphicx}
\usepackage{color, colortbl}
\definecolor{Gray}{gray}{0.9}
\usepackage{subcaption}

\usepackage{stfloats}

\usepackage{microtype}

\usepackage{inconsolata}
\usepackage{amsmath}
\usepackage{cleveref}

\usepackage{enumitem}
\usepackage{makecell}

\graphicspath{ {./images/} }

\date{January 2024}

\begin{document}

    \pagestyle{plain}

    \title{Freely Long-Thinking Transformer (FraiLT)}

    \author{Akbay Tabak \\ {\footnotesize akbay.tabak@frailabs.com } }

    \maketitle

    \begin{abstract}
        Freely Long-Thinking Transformer (FraiLT) is an improved transformer model designed to enhance processing capabilities without scaling up size. It utilizes a recursive approach, iterating over a subset of layers multiple times, and introduces iteration encodings to maintain awareness across these cycles. Iteration encoding allows FraiLT to achieve the interpretive depth of larger models in a compact form. When evaluated on a synthetic story dataset, FraiLT outperformed larger models, showcasing its ability to deliver high-quality performance while reducing memory demands. This model represents a step forward towards more efficient and accessible language models.
    \end{abstract}

    \section{Introduction}

    Advancements in natural language processing (NLP) have been driven by the development of language models that are increasingly complex and capable. However, in the goal of better performance, the computational requirements of these models have also increased, making them impractical for widespread use. Balancing the need for improved capabilities and practicality is a major challenge in the field. FraiLT attempts to solve this problem by optimizing available resources.

    \setlength{\parskip}{1em}

    The conventional approach for processing information in transformer models is linear, where signals are passed sequentially from one layer to another. However, this method requires larger models to perform deeper cognitive processing, which demands more memory. Recently, the Chinchilla\cite{hoffmann2022training} study suggested that models may not be utilizing their full potential, indicating that increased training could result in better utilization of their weights. However, the depth of the model remains a crucial factor\cite{mehta2021delight, mhaskar2016learning, telgarsky2016benefits}, particularly for complex multi-step reasoning tasks. More than extending the training duration for smaller models is required to match the complex reasoning abilities present in larger counterparts.

    \setlength{\parskip}{1em}

    Our brain doesn't linearly process the signals\cite{Galinsky_2020}. Instead, it recursively reutilizes the network. With this inspiration, FraiLT breaks away from the linear approach. It reuses a subset of its layers, iterating the information multiple times, which allows for extended thinking and analysis. The introduction of iteration encodings facilitates this. These learnable parameters inform the model's position within these recursive cycles, enabling nuanced processing that evolves with each iteration.

    \setlength{\parskip}{1em}

    In this article, we explore the architecture and methodology behind FraiLT, detailing how it achieves better performance against larger models on the TinyStories dataset\cite{eldan2023tinystoriesDataset, eldan2023tinystories}, a benchmark tailored for smaller models. We also discuss the implications of FraiLT's design on practical applications, emphasizing its potential to democratize access to high-quality language models by reducing the associated hardware requirements. Through our findings, we aim to contribute to the ongoing discussion on how to create models that are not only powerful but also practical for everyday use.

    \section{Model Architecture}

    FraiLT introduces an innovative decoder-only architecture that builds upon the transformer architecture, as described in ``Attention Is All You Need''\cite{vaswani2023attention} by Vaswani et al. (2017). FraiLT retains the core components of the traditional transformer, including self-attention mechanisms, but diverges in its structural iteration process. The following sections present the unique aspects of the FraiLT architecture.

    \begin{figure*}[ht]
        \centering
        \begin{minipage}[b]{0.30\textwidth}
            \centering
            \vfill
            \includegraphics[width=\textwidth]{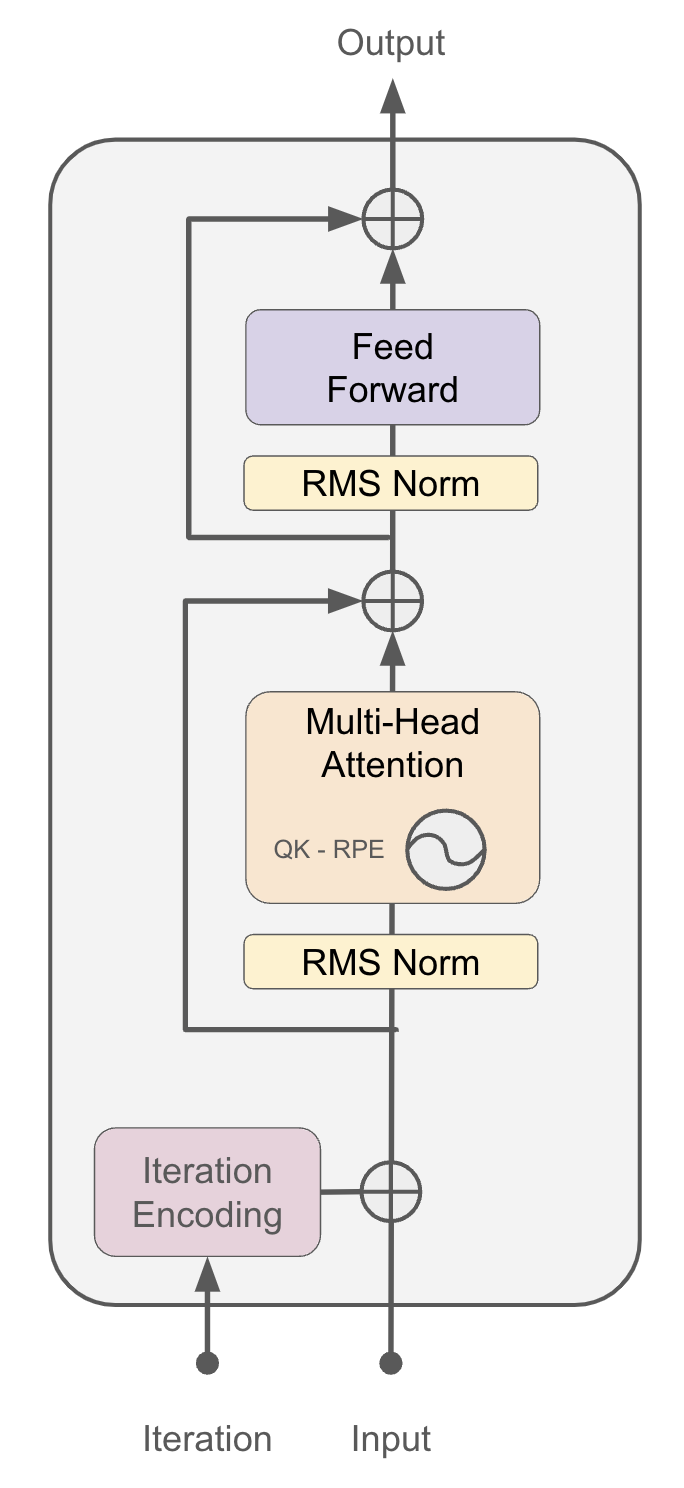}
            \caption{FraiLT Block}
            \label{fig:FraiLT_Block}
        \end{minipage}\hfill
        \begin{minipage}[b]{0.36\textwidth}
            \centering
            \vfill
            \includegraphics[width=\textwidth]{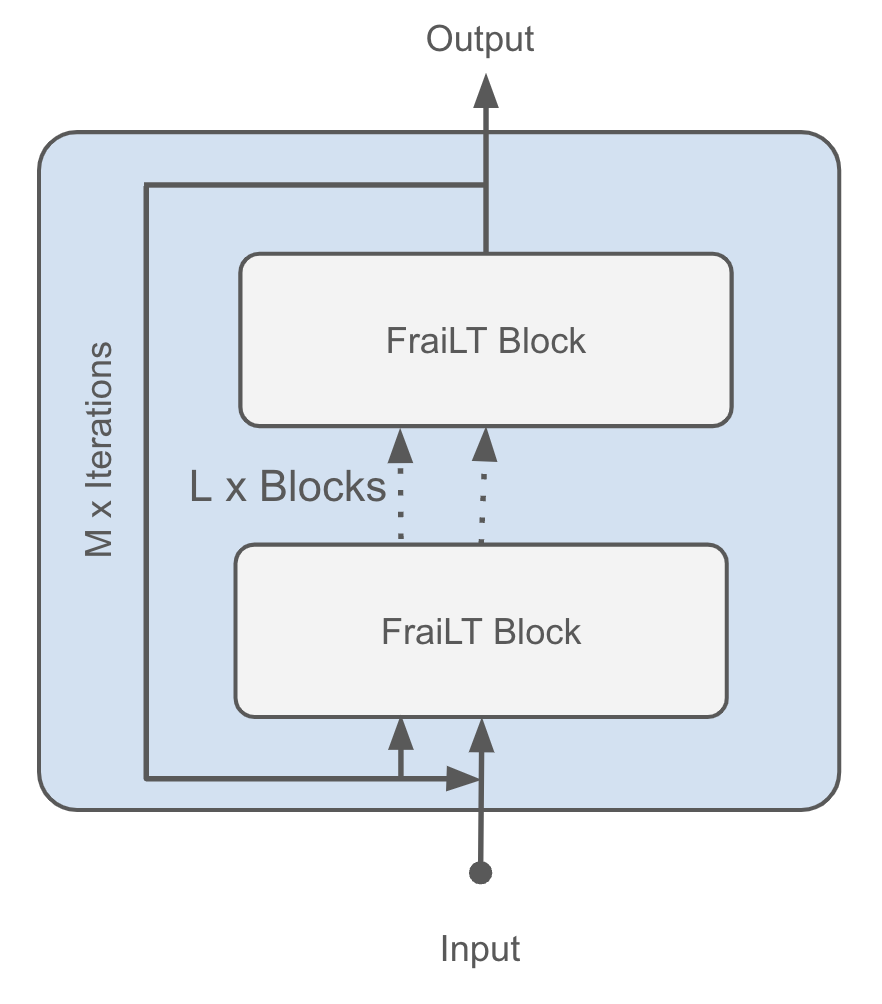}
            \caption{FraiLT Group}
            \label{fig:FraiLT_Block_Group}
        \end{minipage}
        \begin{minipage}[b]{0.33\textwidth}
            \centering
            \vfill
            \includegraphics[width=\textwidth]{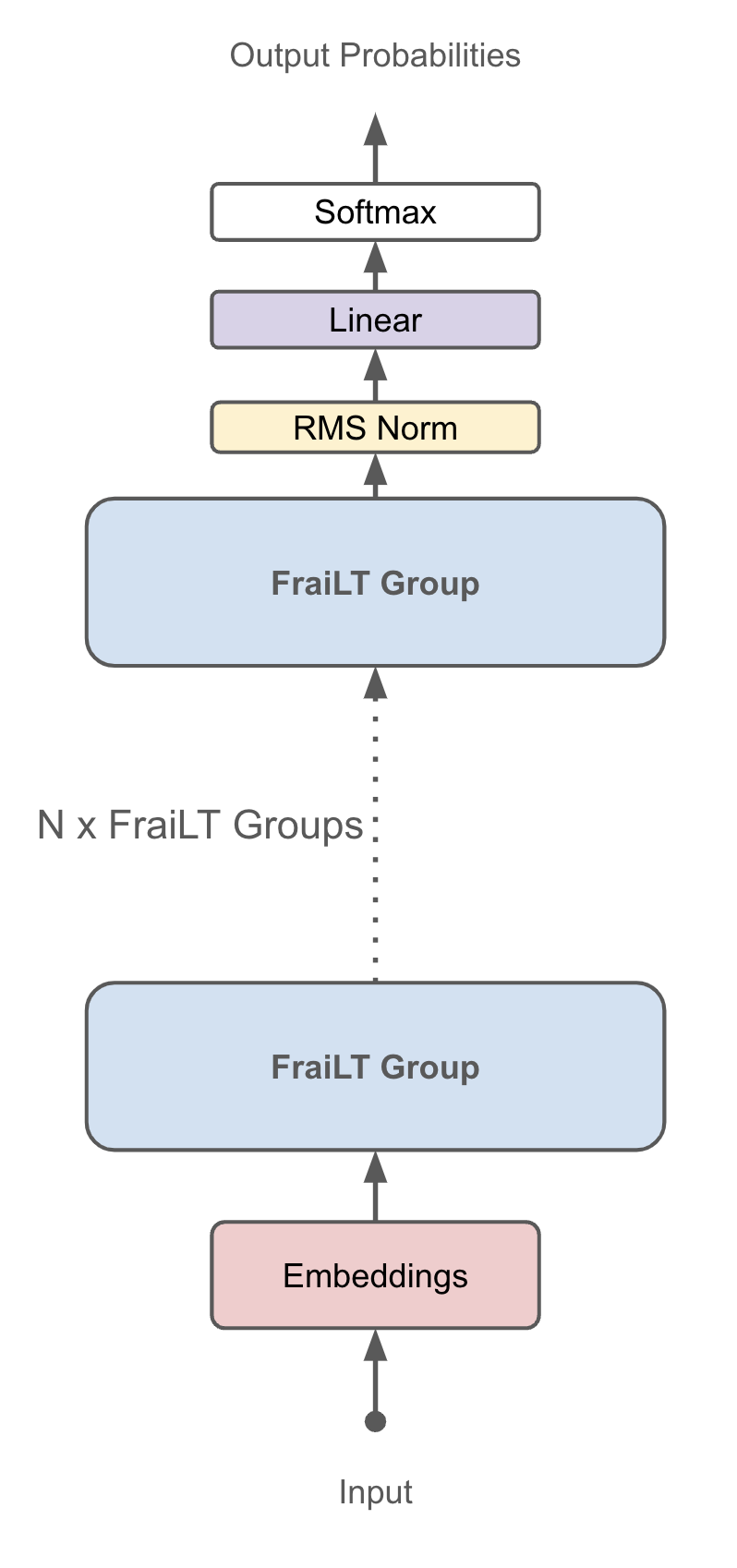}
            \caption{FraiLT Transformer}
            \label{fig:FraiLT_Transformer}
        \end{minipage}
    \end{figure*}

    \paragraph{FraiLT Block:}

    FraiLT Block is an enhanced version of the standard decoder transformer block. Each FraiLT Block is augmented with an iteration-aware mechanism. It receives an iteration number as part of its input, which is then mapped to a corresponding embedding. This iteration embedding is added to the input of the block, allowing the model to modulate its behavior based on the current iteration. This iteration awareness enables the block to adjust its processing strategy dynamically, facilitating deeper and better reasoning with each pass.

    The input tensor $X$ is augmented with an iteration-specific encoding to create an iteration-aware input in the transformer block. For each iteration $i$, an iteration encoding vector $E^{iter}(i)$ is generated, encapsulating the current iteration's information. The augmented input tensor $X_i$ for the $i$-th iteration is obtained by adding the iteration encoding to the original input tensor:

    \[
        X_i = X + E^{iter}(i)
    \]

    Where $X$ is the original input tensor to the transformer block, and $E^{iter}(i)$ represents the iteration encoding for iteration $i$. The function $E^{iter}$ maps each iteration number to a unique encoding vector, enabling the model to modify its processing strategy dynamically with each iteration.

    \paragraph{FraiLT Group:}
    The FraiLT Group is a sequence of \( L \) FraiLT Blocks designed to work together over \( M \) iterations. As information flows through the group, the iteration count increments, and this updated value is fed into each block within the group. The iterative process within a FraiLT Group allows the model to revisit the same sequence of blocks multiple times, expecting to deepen the model's understanding and analysis of the input data.

    Let \( \mathcal{B} = \{B_1, B_2, \ldots, B_L\} \) be the sequence of \( L \) FraiLT Blocks in a FraiLT Group.

    For each iteration \( m \) where \( m = 1, 2, \ldots, M \), and for each block \( B_l \) where \( l = 1, 2, \ldots, L \), the input tensor \( X \) is processed sequentially through the blocks in \( \mathcal{B} \). The output of each block \( B_l \) becomes the input to the next block \( B_{l+1} \), with a unique iteration encoding for each block at each iteration.

    The iterative process can be represented as:

    \[
        X^{(m, l)} = B_l\left(X^{(m, l-1)} + E^{iter}_{l}(m)\right)
    \]

    where:
    \begin{itemize}
        \item \( X^{(m, l)} \) is the output tensor after processing input \( X \) through block \( B_l \) during iteration \( m \).
        \item \( E^{iter}_{l}(m) \) is the iteration encoding for block \( B_l \) at iteration \( m \).
    \end{itemize}

    The final output after \( M \) iterations through the FraiLT Group is \( X^{(M, L)} \), which encapsulates the refined understanding and analysis of the input data through the iterative process across multiple blocks and iterations.

    \paragraph{FraiLT Transformer:}
    The FraiLT Transformer model consists of several components arranged in a specific order. It begins with the standard embedding layer, which processes the input tokens, followed by positional encoding to retain the token order information. The signal then goes through a series of \( N \) FraiLT Groups, each contributing to the iterative refinement of the output. Finally, the model ends with a linear layer and a softmax function, which work together to generate the probability distribution over the output vocabulary.

    In summary, the FraiLT model architecture innovatively extends the capabilities of the transformer by incorporating iteration-aware processing within its blocks and groups. This design enables the model to ``think longer'' over the input and perform complex reasoning tasks typically associated with much larger models, all while maintaining a compact and efficient structure.

    \subsection{Related Work on Weight Sharing in Transformer Models}
    The concept of weight sharing in transformer-based models is not new and has been explored in various architectures to enhance efficiency and performance. This section briefly discusses some of the notable models that have implemented weight-sharing techniques.

    ALBERT\cite{lan2020albert} employs cross-layer parameter sharing and reuses a single transformer block across different layers to reduce the model's size while maintaining competitive performance on language understanding benchmarks.

    Transformer-XL\cite{dai2019transformerxl} introduces segment-level recurrence to learn dependencies beyond fixed lengths, enhancing the model's ability to handle longer sequences and improving performance on language modeling tasks.

    Block-Recurrent Transformers\cite{hutchins2022blockrecurrent} model applies a transformer layer recurrently along a sequence, operating on blocks of tokens to efficiently utilize parallel computation and improve perplexity on language modeling tasks.

    Universal Transformers\cite{dehghani2019universal} combine the parallel processing of feed-forward models with the recurrent structure of RNNs, allowing for improved generalization on tasks that require processing beyond fixed sequence lengths.

    In summary, these models demonstrate the diverse approaches to implementing weight sharing in transformer-based architectures, each with its unique contributions to improving efficiency and performance. The FraiLT model builds upon these ideas, introducing iteration encodings and recursive processing to enhance interpretive depth without scaling up the model size, thereby contributing to the ongoing evolution of efficient language model design.

    \section{Methodology}
    Today's large language models often lead researchers to utilize large training corpus. However, for smaller models, the use of such extensive data can pose challenges. The complexity and diversity of the large corpus may exceed the learning capacity of smaller models, making the training process ineffective. Additionally, using a small portion of a large corpus might not provide a comprehensive representation of the data, leading to skewed learning. Therefore, compact and synthetic data, designed to offer balanced language representation, are more suitable for training smaller models.

    \setlength{\parskip}{1em}

    The synthetic data used in this research was generated by Microsoft Researchers Ronen Eldan and Yuanzhi Li in their work, TinyStories\cite{eldan2023tinystories}. This dataset was carefully designed to provide a balanced representation of simple stories, making it ideal for training smaller models.

    \setlength{\parskip}{1em}

    A comparison model, referred to as the standard model in the rest of this article, was also employed. This model is based on Meta Llama2\cite{touvron2023llama} architecture.

    \subsection{Model Setup}
    We performed a series of experiments to assess the effectiveness of FraiLT in comparison to standard transformer models. The baseline models were selected to represent different sizes, i.e., 1, 2, 4, and 8 layers. These models were used as a control group to evaluate the impact of the iterative approach introduced by FraiLT.

    The standard models were trained on the TinyStories dataset\cite{eldan2023tinystoriesDataset} to establish a performance benchmark. The performance of these models provides a reference point for evaluating the improvements offered by the FraiLT architecture.

    In contrast to the standard models, the FraiLT models were designed to leverage iterative processing within their architecture. To ensure a fair comparison, we ensured that the computational requirements for both the control and test groups were equivalent. This was achieved by adjusting the number of iterations in the FraiLT models to match the computational budget of their standard counterparts. For example, when comparing a 2-layer FraiLT model with an 8-layer standard model, we set the FraiLT model to perform 4 iterations for each of its layers, thus equalizing the computational budget across both models.

    In the FraiLT architecture, it is possible to use different iteration strategies over various subgroups of blocks. However, to maintain simplicity and clarity in this paper, we have presented the results obtained by iterating over individual blocks. This approach has simplified the analysis and provided a clear basis for comparison, demonstrating the effectiveness of the iterative process even when applied to a single block within the FraiLT model.

    This approach allowed us to directly compare the performance of standard models with FraiLT models with fewer layers but more iterations. Our hypothesis was that the iterative process in the FraiLT architecture could compensate for the reduced number of layers and achieve a performance level comparable to that of the standard models with a greater number of layers. The expected performance improvements include not only basic English language understanding but also more complex correlations and analysis, such as consistency throughout the story. This expectation was based on the simulation of deeper networks with iterations.

    \subsection{Training Procedure}

    All models were trained using the same training protocol, optimizer settings, and learning rate schedules to ensure consistency in the experimental conditions. The TinyStories dataset\cite{eldan2023tinystoriesDataset} was used for training, with a held-out portion reserved for validation.

    \setlength{\parskip}{1em}
    The models were trained across various embedding dimensions: 64, 128, 256, 512, and 1024. The specific hyperparameters used for training are shown in Table~\ref{table:model-parameters}.

    \begin{table*}[ht]
        \centering
        \renewcommand{\arraystretch}{1.5}
        \footnotesize
        \begin{tabular}{llp{8cm}}
            \toprule
            \textbf{Name}       & \textbf{Value}              & \textbf{Explanation}                       \\
            \midrule
            Embedding Dimension & 64, 128, 256, 512, and 1024 & Dimensions of the model embeddings         \\
            Layers              & 1, 2, 4, 8                  & Number of transformer layers               \\
            Heads               & 8                           & Number of attention heads                  \\
            Vocabulary Size     & 512                         & Size of the vocabulary used for embeddings \\
            Context Length      & 512                         & Maximum length of input sequences          \\
            \bottomrule
        \end{tabular}
        \caption{Model Parameters}
        \label{table:model-parameters}
    \end{table*}

    \subsection{GPT4-Based Evaluation}

    Our method for evaluating the trained models adopts the GPT-Eval framework as proposed in the TinyStories\cite{eldan2023tinystories} paper, utilizing GPT-4 to assess the quality of the generated texts. This innovative evaluation method extends beyond conventional task-oriented benchmarks, providing a more nuanced assessment that reflects the narrative generation capabilities of language models. By presenting the model with the beginnings of stories and instructing it to craft coherent and creative continuations, we test not only the linguistic accuracy but also the consistency, creativity, and directive-following (plot complication) abilities of the models.

    The process involves feeding the generated story completions, which extend from carefully designed prompts into GPT-4 for analysis. GPT-4 then evaluates these completions, offering both qualitative assessments and quantitative grades in the domains of grammar, creativity, consistency with the story's beginning, and plot coherence. By averaging the scores from multiple completions for each prompt, we obtain a comprehensive view of the performance of FraiLT. This method allows us to draw meaningful comparisons between FraiLT and the standard model.

    We utilized the ``gpt-4-1106-preview'' model from OpenAI in our evaluation approach. Also, to facilitate the extraction of structured scores, we implemented OpenAI's function calling feature. This allowed us to systematically retrieve quantitative evaluations for grammar, creativity, consistency, and plot. Additionally, we updated the prompt from the TinyStories\cite{eldan2023tinystories} paper to better align with the function calling feature. The revised prompt, while maintaining the essence of the original, was rephrased to optimize the clarity and effectiveness of the evaluation process. The exact prompt used in our methodology is as follows:

    \begin{lstlisting}
In the following exercise, the student is given a beginning of a story along with specific instructions for how the story should be completed. The student needs to complete it into a full story that adheres to these instructions. The exercise tests the student's language abilities, creativity, and ability to follow directions. The symbol *** marks the separator between the prescribed beginning and the student's completion.


{story}


Please provide your general assessment about the part written by the student (the one after the *** symbol). Consider the following aspects:


1. Grammar: Is the completion grammatically correct?
2. Creativity: Does the completion show creativity and original thought?
3. Consistency: Is the completion consistent with the beginning of the story?
4. Plot: Does the plot of the completion make sense and is it coherent throughout?


Now, grade the student's completion in terms of the following categories, each on a scale from 1 to 10.
    \end{lstlisting}

    This modification ensured that our instructions to GPT-4 were explicit and tailored to the automated evaluation context, allowing for a more accurate and streamlined assessment of the models.

    \section{Results}

    To evaluate the results of our experiments, we use two primary methods. First, we measure the accuracy of the model with validation loss. Second, we assess the generated text by the model using GPT-4 based on criteria such as grammar, creativity, consistency, and plot. These methods help us to gain a comprehensive understanding of the model's performance.

    \subsection{Validation Loss}

    It is worth noting that although validation loss cannot be the single metric to evaluate the overall performance of FraiLT, it can be used as a reliable indicator for preliminary assessment. The decrease in validation loss shows that the model is continuously learning and improving its prediction capabilities, which indicates the development of its underlying abilities.

    Each model was evaluated across six different embedding layer dimensions: 64, 128, 256, 512, and 1024. The validation loss values for each model across various embedding dimensions are detailed in the following table:

    \begin{table*}[]
        \centering
        \begin{tabular}{lllll}
            \toprule
            Embedding Dim & \multicolumn{4}{c}{Layers} \\
            \cmidrule(lr){2-5}
            & 1-layer & 2-layer & 4-layer & 8-layer \\
            \midrule
            64   & 1.685   & 1.409   & 1.212   & 1.067   \\
            128  & 1.401   & 1.077   & 0.923   & 0.817   \\
            256  & 1.209   & 0.874   & 0.739   & 0.661   \\
            512  & 1.071   & 0.751   & 0.632   & 0.582   \\
            1024 & 0.967   & 0.670   & 0.590   & 0.524   \\
            \bottomrule
        \end{tabular}
        \caption{Validation loss for each model across different embedding dimensions and layer formations for standard models}
        \label{table:embedding-dim-layers}
    \end{table*}

    \begin{table*}[]
        \centering
        \begin{tabular}{lllll}
            \toprule
            Embedding Dim & \multicolumn{4}{c}{Layers} \\
            \cmidrule(lr){2-5}
            & $1^{2}$-layer & $1^{8}$-layer & $2^{4}$-layer & $4^{2}$-layer \\
            \midrule
            64   & 1.586         & 1.522         & 1.348         & 1.190         \\
            128  & 1.215         & 1.136         & 1.007         & 0.895         \\
            256  & 0.969         & 0.887         & 0.791         & 0.716         \\
            512  & 0.796         & 0.712         & 0.648         & 0.601         \\
            1024 & 0.681         & 0.596         & 0.559         & 0.533         \\
            \bottomrule
        \end{tabular}
        \caption{Validation loss for each model across different embedding dimensions and layer formations for FraiLT, where the column titles follow the notation ${layer\_number}^{iteration}$, indicating the number of layers and the number of iterations.}
        \label{table:frailt-embedding-dim-layers}
    \end{table*}

\begin{figure}[H]
        \centering
        \includegraphics[width=0.5\textwidth]{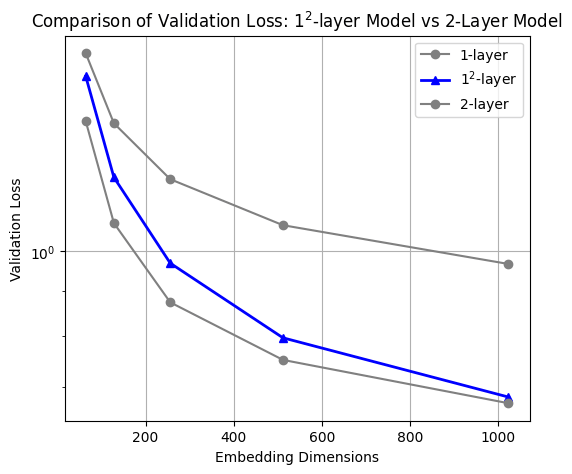}
        \caption{The final validation loss for each model across different embedding dimensions for 1 and 2-layer models}
        \label{fig:validation_loss_1_2_layers}
    \end{figure}

\begin{figure}[H]
        \centering
        \includegraphics[width=0.5\textwidth]{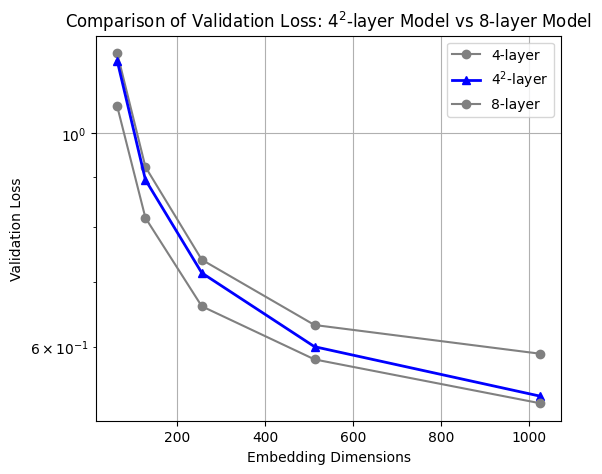}
        \caption{The final validation loss for each model across different embedding dimensions for 4 and 8-layer models}
        \label{fig:validation_loss_4_8_layers}
    \end{figure}

    The corresponding plots visually represent these validation loss values, effectively illustrating the trend as the embedding layer dimensions increase. The plots are shown in Figure~\ref{fig:validation_loss_1_2_layers} and Figure~\ref{fig:validation_loss_4_8_layers}.

    Based on the findings, it was observed that the $1^{2}$ and $4^{2}$-layer FraiLT models showed a consistent reduction in validation loss as the embedding dimension size increased. Remarkably, even at the embedding dimension of 1024, the FraiLT models managed to catch up with the standard models that were double their size (2-layer and 8-layer). This trend was consistent across all tested embedding dimensions, indicating the effectiveness of the FraiLT model.

    \subsection{GPT-4 Evaluation}

    We evaluated generated stories using the GPT-Eval method, which involves inputting the story beginning into the model and generating a completion. Then, GPT-4 evaluates the completion based on grammar, creativity, consistency, and plot coherence. This method offers a more detailed evaluation than traditional methods, using GPT-4's advanced linguistic capabilities to simulate human-like feedback.

    Table~\ref{tab:gpt4-evaluation} shows the GPT-4 evaluation scores for standard and FraiLT models across different categories. The results of the GPT4 evaluation technique reveal insightful data about the performance of FraiLT in comparison to the standard models.

    \begin{table*}[ht]
        \centering
        \includegraphics[width=0.9\textwidth]{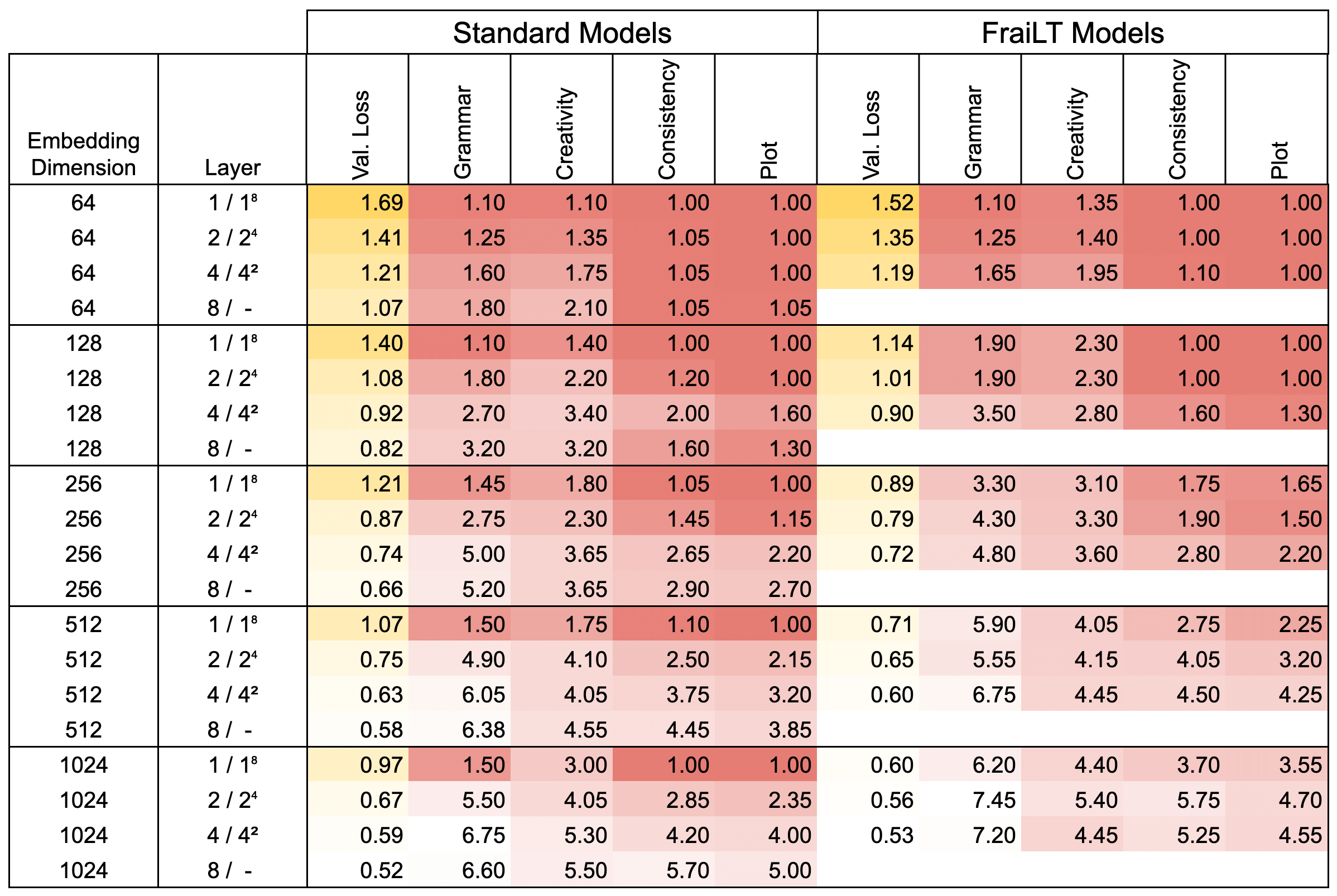}
        \caption{GPT-4 evaluation scores for standard and FraiLT models across different categories. Where the layer numbers follow the notation ${layer\_number}^{iteration}$ for FraiLT models, indicating the number of layers and the number of iterations.}
        \label{tab:gpt4-evaluation}
    \end{table*}

    Table~\ref{tab:gpt4-evaluation-average} presents the average GPT-4 evaluation scores for standard and FraiLT models, highlighting the comparative performance of FraiLT models.

    \begin{table}[H]
        \centering
        \includegraphics[width=0.35\textwidth]{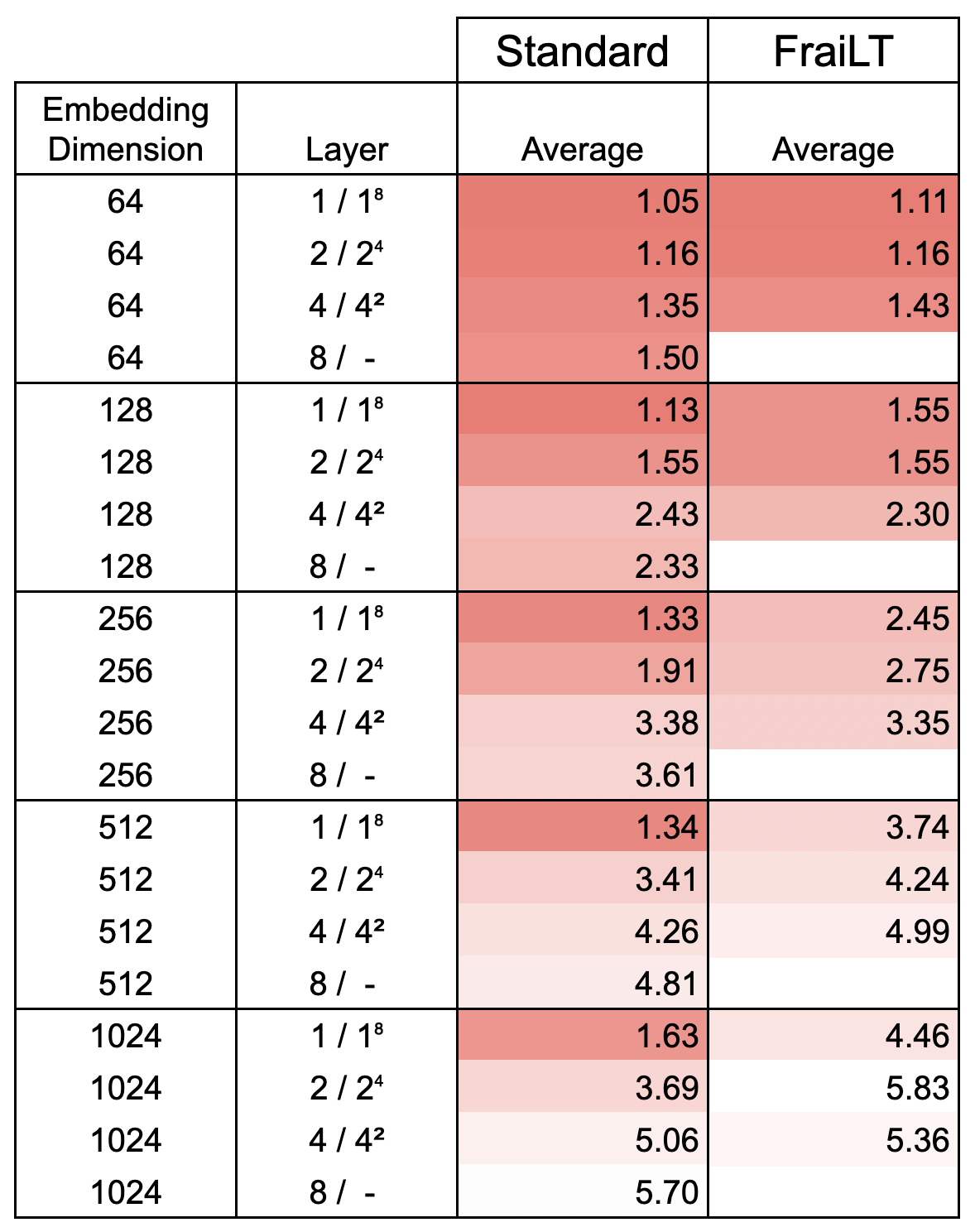}
        \caption{Average GPT-4 evaluation scores for standard and FraiLT models, highlighting the comparative performance. Where the layer numbers follow the notation ${layer\_number}^{iteration}$ for FraiLT models, indicating the number of layers and the number of iterations.}
        \label{tab:gpt4-evaluation-average}
    \end{table}

    It is worth noting that when the validation losses are above 1.0, the models fail to demonstrate any measurable consistency and plot scores according to GPT-4. This trend was observed regardless of the model type. However, this doesn't necessarily mean that the modes lack consistency. Rather, it suggests that GPT-4 does not consider the relationships between the story's start and the complications consistent. We will exclude the region with a validation loss greater than 1.0 to prevent distorted average behavior for the rest of the analysis.

\begin{figure}[H]
        \centering
        \includegraphics[width=0.50\textwidth]{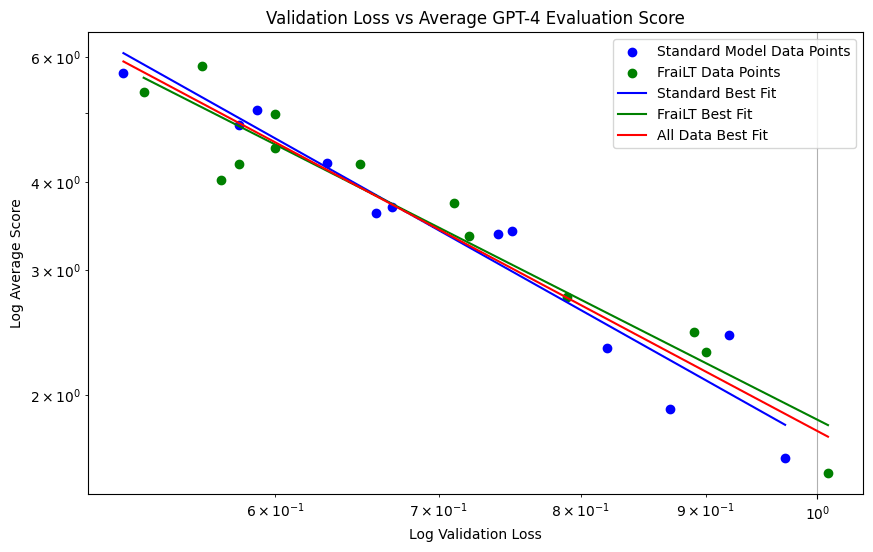}
        \caption{Correlation between average linguistic abilities and validation loss in a logarithmic scale: demonstrating a power law relationship in language model performance for both model types.}
        \label{fig:lossVsAverage}
    \end{figure}

    A linear correlation occurs between the language model's average abilities and validation loss on a logarithmic scale when we exclude high validation loss areas. This correlation follows a predictable power law model, demonstrating how changes to one aspect of a language model affect its overall linguistic capabilities. The relationship is significant and model-type agnostic. Figure~\ref{fig:lossVsAverage} illustrates this finding.

\begin{figure*}[ht]
        \centering
        \includegraphics[width=0.98\textwidth]{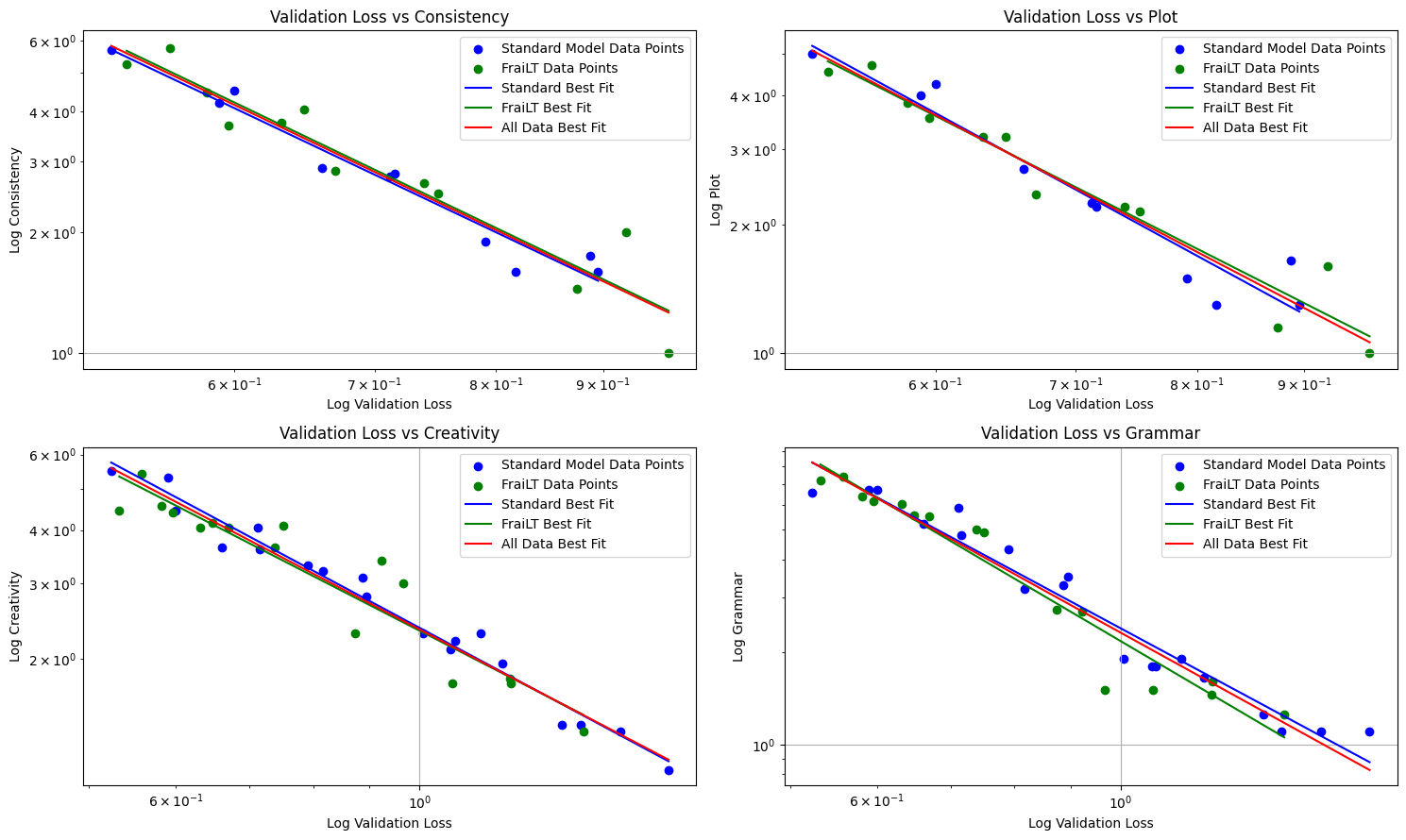}
        \caption{Logarithmic scale analysis of linguistic metrics demonstrating power law relationships for both model types.}
        \label{fig:lossVsMetrics}
    \end{figure*}

    Figure~\ref{fig:lossVsMetrics} provides the logarithmic scale analysis of individual metrics (creativity, grammar, consistency, and plot). Also individual metrics show this power law relationship, which is consistent across both model types.

    In conclusion, the GPT-4 evaluation technique has provided a comprehensive and detailed assessment of our improved models. The findings suggest that the FraiLT models can deliver comparable performance to the standard models across all evaluated categories as long as they can reach the same validation loss. This result meets our initial expectations. Also, the performance results show that the $2^{4}$ and $4^{2}$-layer FraiLT models have demonstrated the ability to reach the performance of the standard 8-layer model across the evaluated categories. This achievement highlights the effectiveness of FraiLT, which has delivered a significant enhancement in model performance without the need for additional layers.

    \section{Discussions \& Conclusions}
    The results of the experiments highlight the effectiveness of our improvements to the standard model. Using validation loss and GPT-4 evaluations for comparison gives us a well-rounded view of the model's performance, covering accuracy as well as language quality.

    FraiLT models consistently displayed superior performance across increasing embedding dimensions, all while maintaining a smaller size. This trend not only emphasizes the successful implementation of the modifications but also suggests that the model's performance may continue to improve with larger embedding sizes.

    Demonstrated findings carry significant implications. The ability of the modified model to deliver high performance with a reduced size could potentially make large language models more accessible, especially on consumer-grade computers. This accessibility could unlock new ways for the deployment of language models. Additionally, the efficiency of the modified models could make them ideal 'student models' in the model distillation process, leading to a substantial reduction in the final model size.

    It is important to keep in mind that the results obtained from these tests are preliminary. Further research and extensive testing are necessary to validate these findings and fully explore the potential of FraiLT on larger model sizes and non-synthetic datasets.

    In conclusion, the FraiLT model has met our initial expectations. The model has demonstrated impressive performance in all domains, similar to that of larger models. This is a promising result in enhancing the performance of language models. These findings lay a strong foundation for future research on FraiLT models, which have the potential to impact the performance of large language models significantly.

    \section{Follow-Up Work}
    Although the initial results are promising, we suggest further work to validate and fully explore the potential of FraiLT.

    \begin{itemize}
        \item \textbf{Evaluation with Larger Datasets:}
        Our models were trained on synthetic data, which, while beneficial for smaller models, may not fully represent the complexity and diversity of natural language. To ensure our model's performance scales with the complexity of the data, we plan to evaluate it on larger and more comprehensive datasets.

        \item \textbf{Training Larger Models:}
        Our initial models were relatively small, which could limit the demonstration of the emerging abilities of language models. We aim to train larger models to identify the phase transition points where these abilities become apparent. This will provide insights into the scalability of our modifications and their impact on more complex models.

        \item \textbf{Exploring Different Iteration Strategies:}
        In this paper, we have presented the results obtained by iterating over individual blocks. However, the FraiLT architecture allows for more complex iteration strategies, which could potentially enhance the model's performance. We plan to explore these strategies to identify the most effective approach for iterative processing.

        \item \textbf{Exploring Different Iteration Encodings:}
        The iteration encodings used in the current model utilized learnable embedding networks. However, this brings additional learning steps during training. We plan to investigate different encoding mechanisms similar to positional encoding to reduce the training time and improve the model's performance.

        \item \textbf{Attention Layers Analysis During Iterations:}
        A deeper understanding of the underlying differences between FraiLT and the standard model can be obtained by probing the attention layers. This will allow us to better understand how the modifications affect the model's learning and prediction capabilities, providing valuable insights for further improvements.
    \end{itemize}

    \onecolumn
    \bibliographystyle{acl_natbib}
    \bibliography{main}

\end{document}